\documentclass[conference]{IEEEtran}
\IEEEoverridecommandlockouts
\usepackage{cite}
\usepackage{amsmath,amssymb,amsfonts}
\usepackage{algorithmic}
\usepackage{graphicx}
\usepackage{textcomp}
\usepackage{xcolor}
\usepackage[color=red]{todonotes}
\usepackage[hidelinks]{hyperref}
\usepackage[T1,T5]{fontenc} 
\DeclareTextFontCommand{\textvietnamese}{\fontencoding{T5}\selectfont}
\def\BibTeX{{\rm B\kern-.05em{\sc i\kern-.025em b}\kern-.08em
    T\kern-.1667em\lower.7ex\hbox{E}\kern-.125emX}}
\begin{document}

\title{BARTPhoBEiT: Pre-trained Sequence-to-Sequence and Image Transformers Models for Vietnamese Visual Question Answering\\}
%
\author{\IEEEauthorblockN{1\textsuperscript{st} Khiem Vinh Tran}
\IEEEauthorblockA{\textit{University of Information Technology}\\
Ho Chi Minh city, Vietnam \\ 
\textit{Vietnam National University}\\
Ho Chi Minh City, Vietnam \\
khiemtv@uit.edu.vn}
\and
\IEEEauthorblockN{2\textsuperscript{nd} Kiet Van Nguyen}
\IEEEauthorblockA{\textit{University of Information Technology}\\
Ho Chi Minh city, Vietnam \\ 
\textit{Vietnam National University}\\
Ho Chi Minh City, Vietnam \\
kietnv@uit.edu.vn}
\and
\IEEEauthorblockN{3\textsuperscript{rd} Ngan Luu Thuy Nguyen}
\IEEEauthorblockA{\textit{University of Information Technology}\\
Ho Chi Minh city, Vietnam \\ 
\textit{Vietnam National University}\\
Ho Chi Minh City, Vietnam \\
ngannlt@uit.edu.vn}
}


\maketitle
\begin{abstract}

Visual Question Answering (VQA) is an intricate and demanding task that integrates natural language processing (NLP) and computer vision (CV), capturing the interest of researchers. The English language, renowned for its wealth of resources, has witnessed notable advancements in both datasets and models designed for VQA. However, there is a lack of models that target specific countries such as Vietnam.
To address this limitation, we introduce a transformer-based Vietnamese model named BARTPhoBEiT. This model includes pre-trained Sequence-to-Sequence and bidirectional encoder representation from Image Transformers in Vietnamese and evaluates Vietnamese VQA datasets.
Experimental results demonstrate that our proposed model outperforms the strong baseline and improves the state-of-the-art in six metrics: Accuracy, Precision, Recall, F1-score, WUPS 0.0, and WUPS 0.9.
\end{abstract}


\section{Introduction}
Multimodal learning, which seeks to bridge vision and language, has gained traction from computer vision and natural language processing communities. Notably, there has been significant progress in various vision-language tasks, including image-text matching, visual captioning, visual grounding, and VQA. However, VQA is a more complex task than other multimodal learning tasks. Recent research has made significant advances in this area.

Visual Question Answering (VQA) is a promising field that combines natural language processing and computer vision to provide accurate answers to questions related to an image. While humans find this task relatively easy, it remains challenging for computers.
VQA demonstrates practical usefulness in various facets of everyday life. Incorporating automated VQA systems in chatbot platforms allows for efficient query response and information retrieval. These systems are essential in several real-world situations, including customer support, recommendation generation, query resolution, dialogic interaction, and customer systems administration. Additionally, VQA exhibits significant potential for informing users of crucial and advantageous information extracted from their surroundings.

In the past few years, there has been a notable surge in research interest regarding Visual Question Answering (VQA) due to the emergence of deep learning techniques and the availability of extensive training datasets. VQA datasets commonly consist of two categories of images: natural images, representing real-world scenarios, and synthetic images, depicting abstract scenes. Moreover, these datasets often encompass two distinct answering modalities: multiple-choice question answering, where the objective is to choose the correct answer from a given set of options, and open-ended question answering, where the goal is to generate an answer without any limitations on the vocabulary employed.

Since 2016, VQA has seen significant advancements due to the development of several benchmarks in resource-rich languages like English. To further enhance the accuracy and efficiency of VQA tasks, researchers have proposed goal-oriented evaluations. However, most studies in this area have focused on languages with abundant resources, neglecting low-resource languages like Vietnamese.

To address this gap, we propose a novel Vietnamese Visual Question Answering method called BARTPhoBEiT, which combines Bidirectional and Auto-Regressive Transformers designed explicitly for the Vietnamese language. We enhance the contextual information captured by the representation model by incorporating the Bidirectional Encoder representation from Image Transformers using Masked Image Modeling (MIM) to recover masked patches and the Vector-Quantized Knowledge Distillation (VQ-KD) algorithm to discretize a semantic space.

The following is how the rest of the article is organized. In \autoref{sec:related}, we review related research on this topic. In \autoref{sec:prob}, we introduce the problem formulation of VQA. Next, \autoref{sec:model} provides a detailed description of our model architecture. We introduce the ViVQA datasets and present our evaluation results in \autoref{sec:ex}. Finally, \autoref{sec:conclude}, we summarize our findings and suggest future directions for research in this area.

\section{Related work}
\label{sec:related}
The VQA task was initially proposed by Antol and colleagues\cite{Antol_2015_ICCV}, where they successfully introduced a new dataset and fundamental methodologies for English. Building upon this success, subsequent studies have expanded to various languages, including Chinese \cite{qi-etal-2022-dureadervis} and Japanese \cite{shimizu-etal-2018-visual}.

Visual Question Answering has received increasing attention from researchers in recent years, with significant advancements in this field. The success of Transformers has been extended from natural language processing to vision \cite{dosovitskiy2021an} and multimodal \cite{pmlr-v139-kim21k} applications, enabling the handling of multiple modalities seamlessly. Various methods have been proposed for applying Transformers in vision-language modeling, depending on the specific nature of downstream tasks. Dual-encoder architectures are used for efficient retrieval \cite{pmlr-v139-radford21a}, ViLT \cite{pmlr-v139-kim21k} utilizes patch embeddings to encode images, which are then concatenated with word embeddings and passed through a Transformer network to learn contextualized representations and model interactions between images and text. ALBEF \cite{li2021align}, on the other hand, uses an Image Transformer \cite{pmlr-v139-touvron21a} to obtain image representations and a text Transformer \cite{devlin-etal-2019-bert} to learn contextualized representations of text.

In recent years, researchers have shown a growing interest in Vietnamese VQA, with various studies exploring its potential and proposing new datasets and methodologies. In 2021, Tran et al. \cite{tran-etal-2021-vivqa} introduced the first ViVQA dataset for VQA in Vietnamese, which has since become a benchmark in subsequent research. Building on this, in 2022, Nguyen et al. \cite{nguyen2023vlsp} proposed a multilingual dataset that includes Vietnamese, further expanding the scope of VQA research in the Vietnamese language.

A variety of new models are regularly being presented to enhance the efficiency of question and answering systems on images. Deeper Long Short Term Memory \cite{Antol_2015_ICCV} and Hierarchical Co-Attention \cite{10.5555/3157096.3157129} are among these models. In Vietnamese, Tran \cite{tran-etal-2021-vivqa} proposed a model using Hierarchical Co-Attention, which is a baseline of ViVQA dataset. 
ViCAN \cite{phucvican} is based on the dense co-attention model proposed by Nam et al. \cite{nam2017dual}. Its objective is to comprehensively understand the interdependencies among the various multimodal features specifically intended for the Vietnamese language. These developments demonstrate the increasing recognition of the importance of VQA in Vietnamese and the efforts being made to advance research in this area.
Recently, with the success of BEIT-1 \cite{bao2022beit} and BEIT-3 \cite{wang2022image} in some A* conferences as ICLR, CVPR in 2022 and 2023. This motivates us to create a new approach for Vietnamese.
\section{Problem formulation}
\label{sec:prob}
The aim of Visual Question Answering is to obtain the most probable answer when given an image and a question by maximizing the conditional probability utilizing a model. Specifically, given an image $I$ and a question $Q$, the task is to find the most probable answer $A$ from a set of answer options $O$. This can be modeled as finding the answer $A$ that maximizes the conditional probability $P(A|I,Q,O)$, that is:
\begin{equation}
A = \operatorname*{argmax}_{A\in O} P(A|I,Q,O)
\end{equation}

where $\operatorname*{argmax}$ is the function that returns the argument that maximizes the given function. The probability $P(A|I,Q,O)$ can be computed using a model that takes the image $I$ and the question $Q$ as inputs and produces the probability distribution over the answer options $O$.
\begin{figure}
    \centering
    \includegraphics[width=0.9\linewidth]{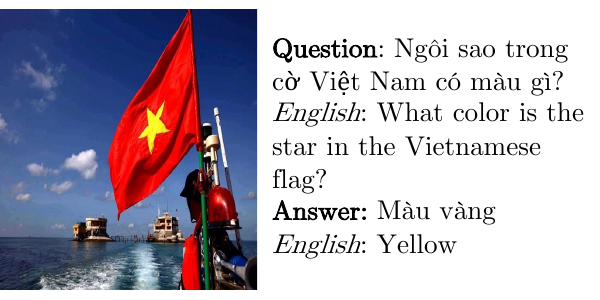}
    \caption{A instance of the visual question
answering task in Vietnamese}
    \label{fig:my_label}
\end{figure}
\section{Our BARTPhoBEiT}
\label{sec:model}
\begin{figure*}
    \centering  \includegraphics[width=0.8\linewidth]{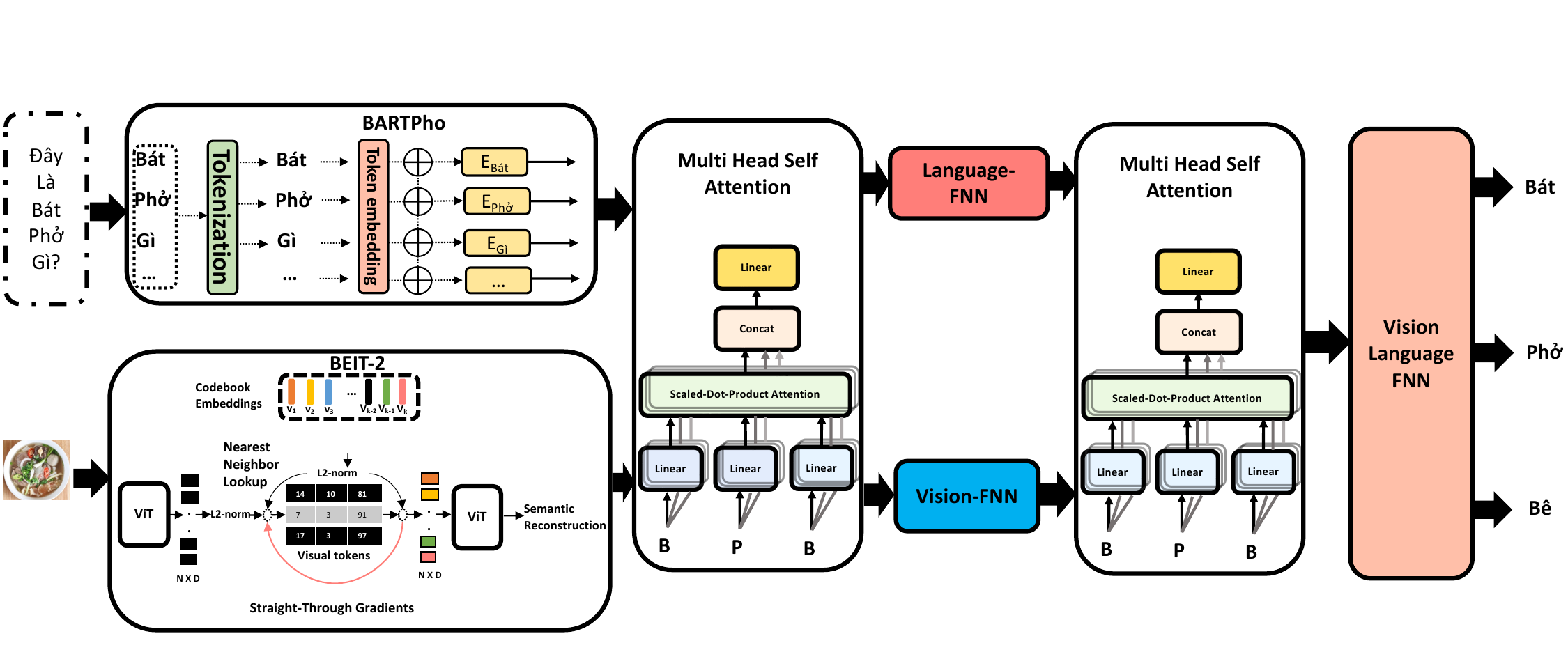}
    \caption{Our core approach using pre-trained sequence-to-sequence
and image transformers for Vietnamese inspired by BEITv3}
    \label{fig:GNNLRP}
\end{figure*}
This section describes the architecture, the pre-training data and
the optimization setup that we use for BARTphoBEiT. Inspired by BEiT v3 \cite{wang2022image}, we construct a new model with the combination of BARTPho \cite{bartpho} for the Vietnamese VQA task.
\subsection{Architecture}
\subsubsection{Multiway Transformers}
The Multiway Transformers \cite{NEURIPS2022_d46662aa} constitute the core of the approach for encoding a range of modalities in our study. Each Multiway Transformer block includes a shared self-attention module and a group of feed-forward networks tailored to process different modalities. During processing, tokens are routed to the relevant experts based on their modality. The model design incorporates both vision and language experts in each layer, with the addition of vision-language experts in the top three layers to support fusion encoders. Incorporating a modality expert pool enhances the capacity of the model to capture modality-specific information. Furthermore, the shared self-attention module can identify correlations between different modalities, facilitating effective fusion for multimodal tasks, such as vision-language processing.
\subsubsection{Masked Data Modeling}
BARTPhoBEiT performs pretraining of the BEIT-3 model using a unified masked data modeling approach \cite{bao2022vl} on both monomodal data  and multimodal data. During pretraining, a random percentage of text tokens or image patches is masked, and the model is trained to recover the masked tokens. This unified mask-then-predict task enables the model not only to learn representations but also to learn the alignment of different modalities. Text data is tokenized and embedded by a BARTPho, whereas image data is tokenized using the BEIT v2 \cite{peng2022beit} to obtain discrete visual tokens as the reconstructed targets.

\subsubsection{BEIT-2} BEIT-2 employs the vision Transformers (ViTs) proposed by Dosovitskiy et al. \cite{dosovitskiy2021an} as the backbone networks to obtain image representations. The input image $x$ is a three-dimensional tensor of size $H \times W \times C$, which is reshaped into $N = \frac{HW}{P^2}$ patches denoted as $\{x_{p}^i\}^N_{i=1}$, where $x_p \in R^{N \times (P^2C)}$ and $(P, P)$ represents the patch size. In experiments, a $224 \times 224$ image is divided into a $14 \times 14$ grid of image patches, each with a size of $16 \times 16$. The image patches $\{x_{p}^i\}^N_{i=1}$ are then flattened and linearly projected to input embeddings for Transformers. The resulting encoding vectors $\{h_i\}^N_{i=1}$ correspond to $N$ image patches. BEIT-2 employs vector-quantized knowledge distillation (VQ-KD) \cite{peng2022beit} to train the visual tokenizer, which maps an image to a sequence of visual tokens or discrete codes. The training objective of VQ-KD is defined as
\begin{equation}
\resizebox{0.5\textwidth}{!}{
    $max \sum_{x \in \mathcal{ D}} \sum_{i=1}^n cos(\boldsymbol{o}_i,\boldsymbol{t}_i) -||sg[\ell_2(\boldsymbol{h}_i)-\ell_2(\boldsymbol{v_{z_i}})]||^2_2 -||\ell_2(\boldsymbol{h}_i) - sg[\ell_2(\boldsymbol{v_{z_i}}]||^2_2$
    },
\end{equation}
where $sg[\cdot]$ stands for the stop-gradient operator, which is an identity at the forward pass while having zero gradients during the backward pass, $\mathcal{D}$ represents the image data used for tokenizer training.
Specifically, an image $x$ is tokenized to $z = [z_1, z_2, \cdots, z_N] \in \mathcal{V}^{(H/P) \times (W/P)}$, where the visual vocabulary (i.e., codebook) $\mathcal{V} \in \mathbb{R}^{K \times D}$ contains $K$ discrete codebook embeddings. The tokenizer consists of a vision Transformer encoder and a quantizer. Firstly, the input image is encoded into vectors by the encoder. Then, the vector quantizer looks up the nearest neighbor in the codebook for each patch representation $h_i$. Let $\{v_1, v_2, \cdots, v_K\}$ denote the codebook embeddings. For the $i$-th image patch, its quantized code is calculated as:
\begin{equation}
\boldsymbol{z}_i = \operatorname*{argmin}_j ||	\ell_2(\boldsymbol{h}_i) - \ell_2(\boldsymbol{v}_j)||_2,
\end{equation}
where $j \in \{1,2,...,K\}$ and $\ell_2$ normalization is used for codebook lookup \cite{yu2022vectorquantized}. The above distance is equivalent to finding codes according to cosine similarity. After the image has been quantized to visual tokens, the $\ell_2$-normalized codebook embeddings $\{\ell_2(\boldsymbol{v}_{z_i})\}^{N}_{i=1}$ are fed into the decoder, which is also implemented as a multi-layer Transformer. The decoder produces output vectors $\{o_i\}^{N}_{i=1}$ that are intended to reconstruct the semantic features of a teacher model. The feature vector of the $i$-th image patch in the teacher model is represented by $t_i$. In the training process, the cosine similarity between the decoder output $o_i$ and the corresponding teacher guidance $t_i$ is maximized.

Since the quantization process is not differentiable, the gradients are directly propagated from the decoder input to the encoder output to enable the back-propagation of gradients to the encoder. In essence, the quantizer performs a lookup of the nearest code for each encoder output, while the gradients of the codebook embeddings provide valuable optimization directions for the encoder.

Specifically, 15\% of tokens from monomodal texts and 50\% of tokens from image-text pairs are randomly masked, while 40\% of image patches are masked using a block-wise masking strategy similar to BEIT \cite{bao2022beit}.
\subsubsection{BARTPho}

BARTPho is implemented using the "large" architecture, which includes 12 encoder and decoder layers, along with the pre-training scheme of BART \cite{lewis-etal-2020-bart}. The pre-training process for BART involves two stages: (i) introducing random noise into the input text using an arbitrary noising function and (ii) training the model to reconstruct the original text by optimizing the cross-entropy between the decoder's output and the original text. BART utilizes the standard Transformer architecture, with the GeLU activation function \cite{hendrycks2016gelu} employed. To perform sentence permutation, BARTPho first groups consecutive sentences to form sentence blocks that are 512 tokens in length. The sentences within each block are then randomly shuffled. Similar to the approach employed in mBART \cite{liu-etal-2020-multilingual-denoising}, this includes a layer-normalization layer on both the encoder and decoder in BARTPho.

We use $BARTpho_\mathrm{word}$ and $BARTpho_\mathrm{syllable}$ for this task, $BARTpho_\mathrm{word}$ leverages the PhoBERT pre-training corpus \cite{nguyen-tuan-nguyen-2020-phobert}, which comprises roughly 20GB of uncompressed text data, containing about 145 million automatically segmented sentences. Additionally, $BARTpho_\mathrm{word}$ uses the same tokenizer employed in PhoBERT, which uses a vocabulary of 64,000 subword types and BPE \cite{sennrich-etal-2016-neural} to segment those automatically segmented sentences into subword units. $BARTpho_\mathrm{word}$ has around 420M parameters and $BARTpho_\mathrm{syllable}$ has
about 396M parameters.
The pre-training data for $BARTpho_\mathrm{syllable}$ is a detokenized variant of the PhoBERT pre-training corpus, i.e., approximately $4$B syllable tokens. For this model, $BARTpho_\mathrm{word}$ and $BARTpho_\mathrm{syllable}$  utilize the pre-trained SentencePiece model \cite{kudo-richardson-2018-sentencepiece} that is implemented in XLM-RoBERTa \cite{conneau-etal-2020-unsupervised} and also used in mBART \cite{liu-etal-2020-multilingual-denoising}, which segments sentences into sub-syllable units and selects a vocabulary of the top 40,000 most frequent types.
\section{Experiments}
\label{sec:ex}
In this section, we perform experiments to assess the performance of BARTPhoBEiT models and previous state-of-the-art (SOTA) models on the ViVQA dataset \cite{tran-etal-2021-vivqa}. We present implementation details and evaluation metrics used to guarantee fairness while comparing these models on the same dataset.
\begin{table*}[!h]
    \caption{The experimental results of different Vietnamese visual question answering systems on ViVQA dataset.}
 \resizebox{\textwidth}{!}{
    \centering

    \begin{tabular}{l|l|l|l|l|l|l}
    \hline
        \textbf{Model} & \textbf{Accuracy} & \textbf{Precision} & \textbf{Recall} & \textbf{F1-score} & \textbf{WUPS 0.0} & \textbf{WUPS 0.9}  \\ \hline
        \\
        $ViHieCoAtt-syllable_{100dims}$  & 0.2954 & 0.2271 & 0.2954 & 0.2234  & 0.9748 & 0.3222 \\ 
        $ViHieCoAtt-word_{100dims}$  & 0.3044 & 0.2049 &  0.3044 & 0.2256  & 0.9761 & 0.3367 \\ 
        $ViHieCoAtt-syllable_{300dims}$  & 0.3014 & 0.2185 & 0.3014 & 0.2277  & 0.9755 & 0.3283 \\ 
        $ViHieCoAtt-word_{300dims}$  & 0.3236 & 0.2192 & 0.3236 & 0.2436  &  0.9751 &  0.3481 \\ 
        \\
        \hline \hline
        \\
        $BARTPhoBeit-base_\mathrm{syllable}$ (Our) & 0.6718 & 0.6684 & 0.6718 & 0.6610 & 0.9751 & 0.6940  \\  
        $BARTPhoBeit-base_\mathrm{word}$ (Our)  &  0.6646 &  0.6593 & 0.6646 & 0.6518 & 0.9758 & 0.6875
        
        \\ 
         $BARTPhoBeit-large_\mathrm{syllable}$ (Our)  & \textbf{0.6858} & \textbf{0.6931} & \textbf{0.6858} & \textbf{0.6777} & \textbf{0.9768} &  \textbf{0.7094}  
        \\
        $BARTPhoBeit-large_\mathrm{word}$  (Our)& 0.6749 &  0.6755 & 0.6749 & 0.6643 & 0.9747 & 0.6950
        \\ \\
       
    \hline
    \end{tabular}}
\label{tab:ex}
\end{table*}
\subsection{Dataset}
The ViVQA dataset comprises 10,328 images and their corresponding 15,000 pairs of questions and answers, which are based on the visual content of the images. The dataset is partitioned into training and test sets using a random allocation approach, with a ratio of 8:2 between the two sets. The dataset categorizes the questions into four distinct types: Object, Number, Color, and Location. These categories account for 41.55\%, 14.81\%, 20.82\%, and 22.82\% of the total questions, respectively.
\subsection{Baseline models}
In this study, we use the Hierarchical Co-Attention Model for Vietnamese, proposed by Tran et al. \cite{tran-etal-2021-vivqa} and serves as the state-of-the-art model in ViVQA, to compare the performance of our proposed model. 

Co-attention is a computational operation that logically utilizes feature information extracted from both an image and a question, allowing for the mutual attention of image and question features. In addition, the hierarchical architecture of co-attention focuses on three types of hierarchies, namely word-image, phrase-image, and question-image, to facilitate a more effective representation of the relationship between the image and the question. The input represents each question as a sequence of T words, with different types of representations at the word, phrase, and question level denoted as $Q^w$, $q_p$, and $q_s$, respectively. Word embeddings are applied to the one-hot encoding vectors of the words in the question, and a 1-D convolution is used to derive the feature representation at the phrase level. Max-pooling layers are then applied to derive phrase-level features, which are subsequently encoded using an LSTM to obtain the question-level representation. The mechanism of Alternating Co-attention operates by iteratively attending to either question or image features based on the guidance of the image or question features.

ViHieCoAtt utilized PhoW2V \cite{tuan-nguyen-etal-2020-pilot} for embedding, which was generated by pre-training a set of 100-dimensional and 300-dimensional syllable embeddings and another set of  100-dimensional and 300-dimensional word embeddings using the Word2Vec skip-gram model \cite{DBLP:journals/corr/abs-1301-3781} on large-scale syllable and word-level corpora consisting of 20GB of Vietnamese texts. Two versions of PhoW2V were created, one with 100-dimensional word embeddings and the other with 300-dimensional word embeddings. Both versions were evaluated in this article to identify the optimal version.


\subsection{Implementation Details}
In our experiments, the following hyper-parameters were used for the model: the training was conducted for 30 epochs with a batch size of 16, a dropout rate of 0.4, weight decay is 0.01, learning rate is 3e-05 and an input image size of 224. The optimizer utilized was Adam. The experiments were performed on a computing system with a CPU of Intel(R) Xeon(R) W-3223 CPU @ 3.50GHz, a GPU of RTX2080 Ti, and a RAM of 64GB.
\subsection{Evaluation metrics}
Prior to analyzing the experimental results, it is crucial to discuss the evaluation metrics utilized to assess the performance of the model. This is consistent with the approach taken by Tran et al. (2018) \cite{tran-etal-2021-vivqa} and Nguyen et al. (2022) \cite{nguyen2023vlsp} in their respective studies. This study employs three performance evaluation metrics, namely F1, Precision, and Recall and Accuracy. The calculation of the F1 score and Accuracy for each answer is based on the tokenized versions of the predicted answer (PA) and the gold answer (GA). The F1 scores for all questions within a given set are then averaged to determine the overall F1 score.
\begin{equation}
     Precision (P) = \frac{GA \cap PA}{PA}
\end{equation}
\begin{equation}
      Recall (R) = \frac{GA \cap PA}{GA}
\end{equation}
\begin{equation}
      F1 = \frac{2 \times P \times R }{P+R}
\end{equation}
\begin{equation}
       Accuracy = \frac{PA}{GA}
\end{equation}

According to
Tran et al. \cite{tran-etal-2021-vivqa}, we also evaluate all models based on WUPS 0.9, and WUPS 0.0. The Wu-Palmer similarity (WUPS) measures semantic similarity between two words based on their classification tree's longest common subsequence. If the similarity score between two words falls below a predefined threshold, the resulting answer may be considered incorrect with a score of 0.
\begin{equation}
    \mathrm{WP}(w_1, w_2) = 2 \times \frac{  \mathrm{depth}(\mathrm{LCS}(w_1, w_2))}{(\mathrm{depth}(w_1) + \mathrm{depth}(w_2))}
\end{equation}

where $\mathrm{LCS}(w_1, w_2)$ is the lowest common subsumer of $w_1$ and $w_2$ in the WordNet hierarchy, and $\mathrm{depth}(w)$ is the depth of the synset corresponding to the word $w$ in the WordNet hierarchy. The WP similarity ranges from 0 (no similarity) to 1 (identical words).

\subsection{Main results}
\autoref{tab:ex} displays the experimental results obtained for SOTA model, HCAN and our four BARTphoBEIT versions on the test sets. The results reveal that all BARTphoBEIT versions significantly outperform the current state-of-the-art (SOTA) across 6 metrics on the test sets.

Our proposed Vietnamese Visual Question Answering (VQA) system demonstrates superior performance according to our experimental findings. In particular, the $BARTPhoBeit-large_\mathrm{syllable}$ version achieves remarkable scores in Accuracy, Precision, Recall, F1-score, WUPS 0.9, and WUPS 0.0, with values of 0.6858, 0.6931, 0.6858, 0.6777, 0.9768, and 0.7094, respectively. Notably, our system outperforms the state-of-the-art (SOTA) in all these metrics. Additionally, our $BARTPhoBeit-large_\mathrm{word}$ version also exhibits superior performance compared to the SOTA in the aforementioned metrics, with scores of 0.6749, 0.6755, 0.6749, 0.6643, 0.9747, and 0.6950, respectively. These results suggest that our VQA system is highly effective in answering questions posed in Vietnamese, with both the word-level and syllable-level versions achieving impressive performance.

Despite BEIT-3 demonstrating promising results and currently being state-of-the-art for the VQAv2 dataset in English, numerous inaccuracies persist in the predicted answers when using it for ViVQA. One of the primary contributing factors to these inaccuracies is the presence of grammatical errors and inaccuracies in translation in ViVQA questions, despite the authors' efforts to identify and rectify erroneous translations using various translation tools, such as Google Translate, Microsoft Translator and double-checking the translations with individuals who achieved high IELTS scores. Nevertheless, the existence of such errors presents a significant challenge in evaluating the similarity between translations, particularly when they do not precisely convey the intended meaning. This issue is especially severe when the data source lacks quality assurance measures, resulting in a large number of errors that may not be easily identifiable. \autoref{tab:error} shows some examples of error sentences in ViVQA.
\begin{table}[!ht]
    \centering
    \caption{Some instances of grammatical errors and inaccuracies in ViVQA}
     \resizebox{0.5\textwidth}{!}{
    \begin{tabular}{l|l}
    \hline
        \textbf{Question} & \textbf{Answer} \\ \hline
        \textvietnamese{Chuối nằm ở đâu?} (Where is the banana?)  & bát (bowl) \\ \hline
        \textvietnamese{Con chó xem cái gì đó đang nấu ăn ở đâu?} \\
        (Where is the dog watching something cooking?)& lò vi sóng (microwave oven) \\ \hline
        \textvietnamese{Ở trên vật nhỏ nào trong lúc ăn} \\ 
        (On what small object while eating) & \textvietnamese{cái ghế} (chair) \\ 
        \hline
        \textvietnamese{Người đàn ông trẻ đang đổ ly rượu ở đâu ở đâu?} \\ 
        (Where is the young man pouring a glass of wine?) & \textvietnamese{phòng bếp} (kitchen) \\ 
        \hline
        \textvietnamese{Phân chia những gì trong sở thú như bao vây?} \\ 
        (What is surrounded by fences in the zoo?) & \textvietnamese{con voi} (elephant) \\ 
        \hline
        \textvietnamese{Cái gì kiểm tra tảng đá trong bao vây của nó?} \\ 
        (What checks its surrounding rock formation?) & \textvietnamese{hươu cao cổ} (giraffe) \\ 
        \hline
        \textvietnamese{Ngồi cạnh nhau với ngọn của họ là gì?} \\ 
        (What is sitting side by side with their tops?) & \textvietnamese{laptop} \\ 
        \hline
        \textvietnamese{Người đàn ông với một con dê và cà vạt giữ} \\ 
        (A man holding a goat and a tie) & \textvietnamese{tách} (glass)\\ 
        \hline
    \end{tabular}} 	
\label{tab:error}
\end{table}
\section{Conclusion and future work}
\label{sec:conclude}
In this study, we introduce BARTphoBEiT, a pre-trained monolingual sequence-to-sequence model and image transformer specifically designed for processing Vietnamese language in the context of visual question answering (VQA). Our research findings demonstrate the superiority of BARTphoBEiT over competing Co-attention models, resulting in state-of-the-art (SOTA) performance for the ViVQA dataset in the Vietnamese language.

While the results are promising, there is always room for improvement. We plan to conduct a deeper linguistic analysis in the future to identify and address the sources of inaccuracies in the system, with the goal of further improving its overall performance. Furthermore, we will enhance the dataset's quality by performing grammatical error checks, thus ensuring that the questions are properly formulated and accurately reflect the intended meaning. Ultimately, we believe these efforts will help advance the field of VQA and contribute to more effective and reliable natural language processing in the Vietnamese language.

\section*{Acknowledgment}
This research is funded by University of Information Technology, Vietnam National University HoChiMinh City under grant number D1-2023-15

\bibliographystyle{IEEEtran}
\bibliography{ref.bib}{}

\end{document}